\documentclass[twocolumn,
]{ceurart}

\usepackage{avm}

\usepackage{url}

\usepackage{gb4e}

\usepackage{forest,adjustbox}
\useforestlibrary{linguistics}
\forestapplylibrarydefaults{linguistics}

\begin{document}

\copyrightyear{2024}
\copyrightclause{Copyright for this paper by its authors.
  Use permitted under Creative Commons License Attribution 4.0
  International (CC BY 4.0).}

\conference{SEPLN-CEDI2024: Seminar of the Spanish Society for Natural Language Processing at the 7\textsuperscript{th} Spanish Conference on Informatics, June 19-20, 2024, A Coruña, Spain}

\title{Grammar Assistance Using Syntactic Structures (GAUSS)}

\author[1]{Olga Zamaraeva}[%
orcid=0000-0001-9969-058X,
email=olga.zamaraeva@udc.es,
url=https://olzama.github.io/,
]
\address[1]{Universidade da Coruña, CITIC, Department of Computer Science and Information Technologies. 15071 A Coruña, Spain}

\author[2]{Lorena S.\ Allegue}[%
orcid=0009-0009-5529-4150,
email=l.sallegue@udc.es,
]
\address[2]{Universidade da Coruña, CITIC, Department of Humanities (``Letras''). 15071 A Coruña, Spain}

\author[1]{Carlos Gómez-Rodríguez}[%
orcid=0000-0003-0752-8812,
email=carlos.gomez@udc.es,
url=https://www.grupolys.org/~cgomez,
]

\author[2]{Margarita Alonso-Ramos}[%
orcid=0000-0002-1353-9270,
email=margarita.alonso@udc.es,
]

\author[3]{Anastasiia Ogneva}[%
orcid=0000-0003-0237-7146,
email=anastasiia.ogneva@usc.es,
url=https://sites.google.com/view/anastasiiaogneva,
]
\address[3]{Universidade de Santiago de Compostela, Department of Developmental Psychology, 15782 Santiago de Compostela, Spain}

\begin{abstract}
 Automatic grammar coaching serves an important purpose of advising on standard grammar varieties while not imposing social pressures or reinforcing established social roles. Such systems already exist but most of them are for English and few of them offer meaningful feedback. Furthermore, they typically rely completely on neural methods and require huge computational resources which most of the world cannot afford. We propose a grammar coaching system for Spanish that relies on (i) a rich linguistic formalism capable of giving informative feedback; and (ii) a faster parsing algorithm which makes using this formalism practical in a real-world application. The approach is feasible for any language for which there is a computerized grammar and is less reliant on expensive and environmentally costly neural methods. We seek to contribute to Greener AI and to address global education challenges by raising the standards of inclusivity and engagement in grammar coaching.
\end{abstract}


\maketitle

\section{Introduction}

The GAUSS project is concerned with \textbf{a new, faster parsing technology for grammar coaching} and will develop a Spanish grammar coaching system. Automatic grammar coaching helps people write more like a native speaker of a language would, thus helping them navigate around biases associated with language. This is important for (i) finding a job and counterbalancing latent discrimination in any given society, in the case of major languages like Spanish; and (ii) reinforcing the understanding that each language has a systematic grammar in its own right, in the case of minority languages (like e.g.\ Galician). Grammar coaching systems rely on parsing to determine (i) that grammar in a sentence could be improved; and (ii) how specifically to improve it. Parsing is mapping a sentence to a structure (Figure 1).
 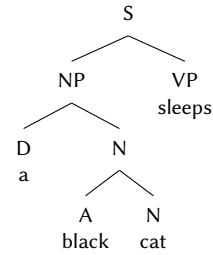
\begin{figure}[h!]
 \centering
 \begin{forest}
 [S[NP[D\\a][N[A\\black][N\\cat]]][VP\\sleeps]]
 \end{forest}
      \caption{A simplified syntactic parse.}
\end{figure}

The project uses an implemented linguistic grammar of Spanish to provide meaningful feedback on writing. The notion of grammaticality encoded in such grammars is more descriptive than prescriptive; the system will not try reinforce someone's opinion on what is correct and what is not.  Our specific contribution will be in (i) developing such a system for Spanish, one of the world's most spoken languages, leveraging an existing body of linguistic knowledge; and (ii) making the underlying parsing technology fast enough to be deployed at scale. A Spanish system based on cross-linguistically applicable methodology will pave the way for other European languages including minority languages, starting from Galician, the language of our host province. The main challenge we will address is integrating neural and symbolic approaches to parsing, demonstrating that expensive neural methods can be applied in a limited manner, and that the computational ``price tag'' of NLP technology can be reduced.

\section{State of the art at the start of the project}
Most grammar coaching systems available today are purely statistical and do not use explicit linguistic knowledge. Based on purely statistical methods and lacking interpretability, they ``guess'' based on the context and are not aware of concepts like agreement. Their feedback is divorced from the methodology of suggesting a better sentence, opening possibilities for wrong feedback. Such systems are often only available for English, because their neural architectures require huge quantities of training data. Such systems are also ecologically problematic\citep{schwartz2020green}.

\section{Methodology}
The GAUSS project is the result of the collaboration between research areas such as CS, NLP, theoretical linguistics, and applied linguistics. The intersectional nature of the project is realized by the combination of NLP techniques and theoretically formalized grammars. In particular, the project relies on the Spanish Resource Grammar \citep[SRG;][]{marimon2010spanish, marimon2014automatic, zam:srg:2024}, a grammar of Spanish implemented in the Head-driven Phrase Structure Grammar formalism (HPSG).

\subsection{HPSG syntax theory}
\begin{figure*}[h!]
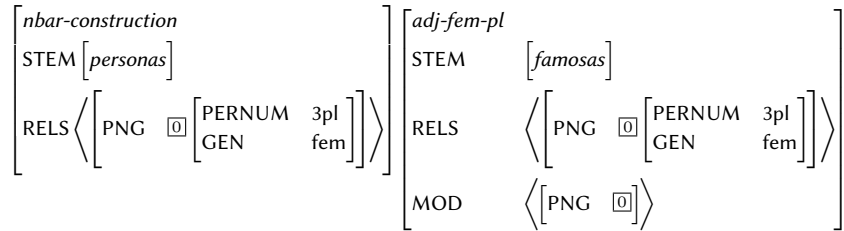

    \centering
\begin{avm}
\[\textit{nbar-construction} \\ 
STEM  \[ \emph{personas} \]  \\ 
RELS  \< \[ PNG & \@0 \[ PERNUM & 3pl \\ GEN & fem \] \]\> \]
\end{avm}
\begin{avm}
\[\textit{adj-fem-pl} \\
STEM & \[ \emph{famosas} \]  \\
RELS & \< \[ PNG & \@0 \[ PERNUM & 3pl \\ GEN & fem \] \]\> \\
MOD & \< \[ PNG & \@0 \] \> 
 \]
\end{avm}
    \caption{Two simplified HPSG structures that can form a phrase `famous persons' in Spanish, \textit{personas famosas}.}
    \label{fig:feature-structures}
\end{figure*}

Head-driven Phrase Structure Grammar \citep[HPSG;][]{Pol:Sag:94} is a constraint unification theory of syntax. A sentence is analyzed as a structure where parts can be constrained to be identical to each other. For example, a verb's agreement values (e.g.\ third person) can be constrained to be identical to the agreement values of the subject of the verb. Similarly, adjectives can be constrained with respect to the agreement values of the noun they modify, as shown in Figure \ref{fig:feature-structures}. Crucially, ungrammatical strings of words will violate the constraints required for well-formed structures and as such will not be covered by an HPSG grammar.

Structures like the ones in Figure \ref{fig:feature-structures} are instances of more general types and can be seen in the specific results of deploying the grammar on some data. The grammar itself contains the types, not the instances. The types are instantiated through interfacing with the lexicon and, in some cases, an external morphophonological analyzer.

The HPSG theory covers many syntactic phenomena and has been developed and tested using a variety of data from a variety of languages. One of the approaches to the empirical testing of this theory is implementing it on the computer and then automatically parsing data and inspecting the results for correctness and consistency. Efforts of this kind include ParGram \citep{butt2002urdu}, CoreGram \citep{muller2015coregram} and DELPH-IN \citep{copestake2000appendix, bender:emerson:handbook} It is this approach that gave rise to the SRG.

\subsection{DELPH-IN Consortium}
The DELPH-IN research consortium is an international effort for grammar engineering using HPSG: Deep Linguistic Processing with HPSG Initiative. It is committed to using a particular version of the HPSG formalism that was defined originally in \citep{copestake2000appendix}. The consortium develops tools such as parsers, including the parser we used in this project, the ACE parser \citep{crysmann2012towards}. Another set of relevant tools includes the software for automatic profiling of test data known as \texttt{incr tsdb()} (pronounced `tsdb++') \citep{oepen1998towards, oepen1999incr} and a related tool ``full-forest treebanker'' (fftb) \citep{packard2014uw}. These tools allow us to inspect differences between different grammar versions systematically.

\begin{figure*}
\centering
\includegraphics[width=\textwidth]{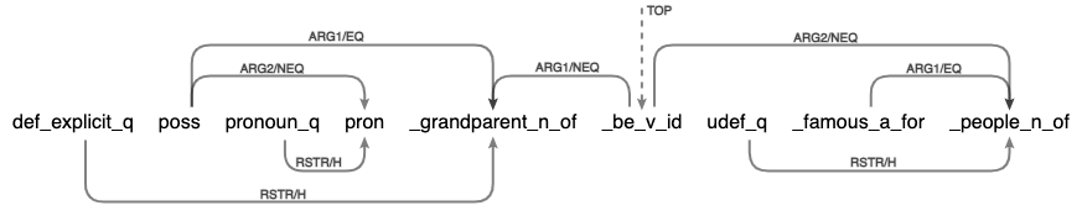}
\caption{Dependency structure for ``My grandparents are famous people''.}
\label{fig:dmrs}
\end{figure*}

Grammars are tested on sentences automatically, using a parser. The first time a grammar is run on a sentence, an expert must verify the correctness of the output. Often it makes sense to do this by looking at the semantic (dependency) structure; we can assume that if the semantics is correct, then the syntactic structure that corresponds to it is adequate. The semantics in DELPH-IN grammars is modeled with Minimal Recursion Semantics formalism \citep[MRS;][]{copestake2005minimal}. An MRS structure is a bag of predications encoding dependencies as well as modifier and negation scope, information structure, and more. It can be automatically converted to a dependency structure familiar to natural language processing (NLP) practitioners (Figure \ref{fig:dmrs}). When the parser analyzes a sentence according to the grammar, the resulting structure includes an MRS, the adequacy of which is easy to establish manually (whether the meaning of the sentence is the intended one). Adequacy of obtained analyses on corpora serve as accumulating evidence for the validity of the theory of syntax.

\subsection{Spanish Resource Grammar}
At the core of the project's methodology is the digital representation of the Spanish syntax, the Spanish Resource Grammar \citep{marimon2010spanish, marimon2014automatic, zam:srg:2024}. The SRG consists of 54,510 lemmas in the lexicon, 543 lexical types to instantiate those lemmas, 504 lexical rule types serving morphophonological analysis, and 226 phrasal types. It is the second largest DELPH-IN grammar (after the English Resource Grammar \citep{flickinger2000building, Flickinger:11}). SRG was first developed prior to the ACE parser and one of the objectives of the GAUSS project ended up being the complete reimplementation of the SRG morphophonological interface. The outcome is that the SRG can now be used with the ACE parser \citep{zam:srg:2024}. As before, it relies on an external morphophonological analyzer Freeling \citep{carreras2004freeling}.

One major outcome of this is that we could reparse the portions of the AnCora corpus previously released as the TIBIDABO treebank \citep{marimon2014automatic}. The previously released version was partially verified for the correctness of the structure but the accuracy figures corresponding to that verification were never reported (as far as we can tell). One of the outcomes of GAUSS is the re-parserd, re-verified, and re-released portions of TIBIDABO (currently 2291 sentences) \citep{zam:srg:2024}. The updated version of the SRG along with the verified treebanks are open-source and are released on GitHub: \url{https://github.com/delph-in/srg}

\subsection{Using the SRG with learner data}
The main idea behind the GAUSS project is that we can use the SRG to model constructions characteristic of learners of Spanish (as opposed to native speakers). We create a version of the SRG that is modified specifically to cover learner constructions, starting with gender agreement constructions, like the one illustrated in example (\ref{ex:spa1}). 

\begin{exe}
\ex \gll *Mis abuelos son personas famosos.\\
my.{\sc 3pl} grandparent.{\sc masc.pl} be.{\sc 3pl.pres.ind} person.{\sc fem.3pl} famous.{\sc masc.pl}\\
Intended: `My grandparents are famous people.' [spa; \citealt{yamada2020cows}]
\label{ex:spa1}
\end{exe}

\begin{figure*}[h!]
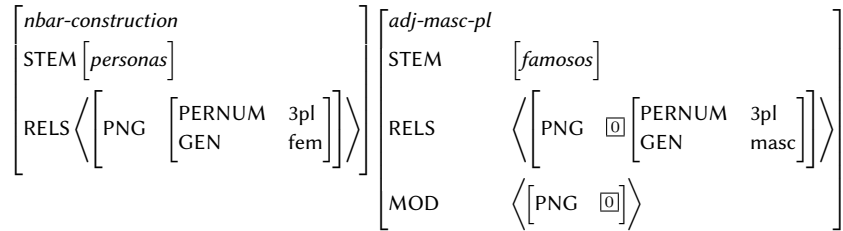

    \centering
\begin{avm}
\[\textit{nbar-construction} \\ 
STEM  \[ \emph{personas} \]  \\ 
RELS  \< \[ PNG & \[ PERNUM & 3pl \\ GEN & fem \] \]\> \]
\end{avm}
\begin{avm}
\[\textit{adj-masc-pl} \\
STEM & \[ \emph{famosos} \]  \\
RELS & \< \[ PNG & \@0 \[ PERNUM & 3pl \\ GEN & masc \] \]\> \\
MOD & \< \[ PNG & \@0 \] \> 
 \]
\end{avm}
    \caption{HPSG feature structures representing incompatible agreement values in the learner production \textit{personas famosos}.}
    \label{fig:agr}
\end{figure*}

The grammar will detect such learner structures using what is called `mal-rules' \citep{schneider1998recognizing}, a technical term for HPSG types designed specifically to cover productions characteristic of learners. For example, the grammar will have to have a way to ignore the incompatible agreement values in Figure \ref{fig:agr}.

We achieve this by only a small set of modifications to the grammar. We use the interface of the grammar with the external morphophonological analyzer to recognize any noun or adjective as potentially belonging to either gender (this requires 40  short additional entries in the lexical rule section of the grammar, one corresponding to each possible Freeling noun or adjective tag). We associate each such lexical rule with a special LEARNER feature, so that ultimately any sentence that uses one or more of such rules can be detected as a learner production. No changes in the syntax part of the grammar are required, in principle. However, deploying the grammar on the learner sentences without modifications revealed a number of overgeneration issues in the original grammar, which we were able to fix thanks to this experiment. Overgeneration is when a grammar covers an ungrammatical sentence or produces a nonsensical structure for a sentence along with the correct one(s). When we saw instances of the original grammar covering learner productions, we investigated such cases and have found 4 syntactic types (so far) which were underconstrained with respect to the agreement values. We have added the missing agreement constraints, which resulted in reduced overgeneration and ambiguity of the SRG with respect to the TIBIDABO treebank. In this way, modeling learner constructions helped us improve the analysis of agreement in the original SRG.

After all the necessary mal-rules are implemented, the plan is to (1) accompany each model of a learner construction with meaningful feedback; and (2) deploy the grammar as a web-based service such that it can be tested by learners of Spanish. This is work in progress.

\subsection{Parsing speed bottleneck}
The main challenge in HPSG parsing speed is that large feature structures combinatorically lead to huge search space. As a result, HPSG parsing is comparatively slow in practice. For example, the ACE parser takes about 3 seconds per sentence on average on a corpus of 100K sentences (some of these sentences take minutes while others take less than a second) \citep{zamaraeva2023revisiting}. The GAUSS project attempts to address this challenge by a combination of methodologies: (1) improving analyses in the grammar to reduce meaningless ambiguity (overgeneration) and thus reduce the size of the parse chart; (2) integrating top-down parsing, and (3) filtering lexical entries and grammar rules so that fewer rules are considered at each step. Method (1) is what we employed while addressing overgeneration we discovered by deploying the grammar on the learner corpus. We have managed to improve the SRG's performance up to 60\% on sentences of length 8-10. Method (2) has been underexplored in HPSG but has seen a rekindled interest recently \citep{chiruzzo2020statistical}. HPSG parers are overwhelmingly bottom-up but for long sentences, a lot can be learned immediately from the start of the sentence/top of the syntax tree, discarding many irrelevant search paths. Method (3) includes developing a neural supertagger (filter) for HPSG. The supertagger will reduce the number of possibilities the parser needs to explore by discarding unlikely word meanings. Statistical filtering was successfully applied to HPSG \citep{dridan2013ubertagging}, and we are now researching how neural methods can improve the SOTA. We start with applying method (3) to the English Resource Grammar treebanks and obtain a speed-up of a factor of three compared to the baseline. However, when we attempted the method on the Spanish treebanks, the results were not yet satisfactory, apparently because the Spanish treebanks were not big enough at the start of the GAUSS project. Now that we added more verified items in the treebanks, we can attempt to train a neural supertagger for Spanish once again.

\section{Planning and Team}

The GAUSS project consists of three Research Objective (RO) and four Work Packages (WP). They are summarized in Table \ref{tab:RO}.

\begin{table}[h!]
\caption{Research Objectives (RO).}
    \begin{tabular}{lll}
         \textbf{RO} & \textbf{WP} & \textbf{Objective} \\
         \hline
         RO1 & WP1 & Fast HPSG parsing   \\ 
         RO2 & WP2 & Spanish error productions in HPSG  \\
         RO3 & WP3--4 & Empirical integration of RO1--2 
    \end{tabular}
    
    \label{tab:RO}
\end{table}

The team consists of the PI MSCA postdoctoral fellow Olga Zamaraeva, supervisor Carlos Gómez-Rodríguez, co-supervisor Margarita Alonso-Ramos, collaborator Anastasiia Ogneva, and research assistant Lorena S.\ Allegue. Olga Zamaraeva does most of the technical and organizational work. Lorena S.\ Allegue verifies the correctness of the grammar output. Carlos Gómez-Rodríguez advises on computational issues. Margarita Alonso-Ramos advises on the use of the learner corpora. Anastasiia Ogneva advises on second language acquisition theory.

\begin{acknowledgments}
 The GAUSS project is funded by the European Union's Horizon Europe Framework Programme under the Marie Skłodowska-Curie postdoctoral fellowship grant HORIZON-MSCA‐2021‐PF‐01 (GAUSS, grant agreement No 101063104) The project is carried out in the Language and Society Information research group (LyS) of Universidade da Coruña. 
 \end{acknowledgments}

\bibliography{master}

\begin{thebibliography}{22}
\expandafter\ifx\csname natexlab\endcsname\relax\def\natexlab#1{#1}\fi
\providecommand{\url}[1]{\texttt{#1}}
\providecommand{\href}[2]{#2}
\providecommand{\path}[1]{#1}
\providecommand{\DOIprefix}{doi:}
\providecommand{\ArXivprefix}{arXiv:}
\providecommand{\URLprefix}{URL: }
\providecommand{\Pubmedprefix}{pmid:}
\providecommand{\doi}[1]{\href{http://dx.doi.org/#1}{\path{#1}}}
\providecommand{\Pubmed}[1]{\href{pmid:#1}{\path{#1}}}
\providecommand{\bibinfo}[2]{#2}
\ifx\xfnm\relax \def\xfnm[#1]{\unskip,\space#1}\fi
\bibitem[{Schwartz et~al.(2020)Schwartz, Dodge, Smith, and Etzioni}]{schwartz2020green}
\bibinfo{author}{R.~Schwartz}, \bibinfo{author}{J.~Dodge}, \bibinfo{author}{N.~A. Smith}, \bibinfo{author}{O.~Etzioni},
\newblock \bibinfo{title}{Green {AI}},
\newblock \bibinfo{journal}{ACM} \bibinfo{volume}{63} (\bibinfo{year}{2020}).
\bibitem[{Marimon(2010)}]{marimon2010spanish}
\bibinfo{author}{M.~Marimon},
\newblock \bibinfo{title}{The {S}panish {R}esource {G}rammar},
\newblock in: \bibinfo{booktitle}{LREC}, \bibinfo{year}{2010}.
\bibitem[{Marimon et~al.(2014)Marimon, Bel, and Padr\'{o}}]{marimon2014automatic}
\bibinfo{author}{M.~Marimon}, \bibinfo{author}{N.~Bel}, \bibinfo{author}{L.~Padr\'{o}},
\newblock \bibinfo{title}{Automatic selection of {HPSG}-parsed sentences for treebank construction},
\newblock \bibinfo{journal}{Computational Linguistics} \bibinfo{volume}{40} (\bibinfo{year}{2014}) \bibinfo{pages}{523--531}.
\bibitem[{Zamaraeva et~al.(ress)Zamaraeva, S.~Allegue, and Gómez-Rodríguez}]{zam:srg:2024}
\bibinfo{author}{O.~Zamaraeva}, \bibinfo{author}{L.~S.~Allegue}, \bibinfo{author}{C.~Gómez-Rodríguez},
\newblock \bibinfo{title}{Spanish {R}esource {G}rammar version 2023},
\newblock in: \bibinfo{booktitle}{COLING-2024}, \bibinfo{year}{in press}.
\bibitem[{Pollard and Sag(1994)}]{Pol:Sag:94}
\bibinfo{author}{C.~Pollard}, \bibinfo{author}{I.~Sag}, \bibinfo{title}{Head-{D}riven {P}hrase {S}tructure {G}rammar}, \bibinfo{publisher}{CSLI}, \bibinfo{address}{Stanford, CA}, \bibinfo{year}{1994}.
\bibitem[{Butt and King(2002)}]{butt2002urdu}
\bibinfo{author}{M.~Butt}, \bibinfo{author}{T.~H. King},
\newblock \bibinfo{title}{Urdu and the parallel grammar project},
\newblock in: \bibinfo{booktitle}{Proceedings of the 3rd workshop on Asian language resources and international standardization-Volume 12}, \bibinfo{organization}{Association for Computational Linguistics}, \bibinfo{year}{2002}, pp. \bibinfo{pages}{1--3}.
\bibitem[{M{\"u}ller(2015)}]{muller2015coregram}
\bibinfo{author}{S.~M{\"u}ller},
\newblock \bibinfo{title}{The {C}ore{G}ram project: Theoretical linguistics, theory development and verification},
\newblock \bibinfo{journal}{Journal of Language Modelling} \bibinfo{volume}{3} (\bibinfo{year}{2015}) \bibinfo{pages}{21--86}.
\bibitem[{Copestake(2000)}]{copestake2000appendix}
\bibinfo{author}{A.~Copestake},
\newblock \bibinfo{title}{Appendix: Definitions of typed feature structures},
\newblock \bibinfo{journal}{Natural Language Engineering} \bibinfo{volume}{6} (\bibinfo{year}{2000}) \bibinfo{pages}{109--112}.
\bibitem[{Bender and Emerson(2021)}]{bender:emerson:handbook}
\bibinfo{author}{E.~M. Bender}, \bibinfo{author}{G.~Emerson},
\newblock \bibinfo{title}{Computational linguistics and grammar engineering},
\newblock in: \bibinfo{editor}{S.~Müller}, \bibinfo{editor}{A.~Abeillé}, \bibinfo{editor}{R.~D. Borsley}, \bibinfo{editor}{J.-P. Koenig} (Eds.), \bibinfo{booktitle}{Head-Driven Phrase Structure Grammar: The handbook}, \bibinfo{year}{2021}.
\bibitem[{Crysmann and Packard(2012)}]{crysmann2012towards}
\bibinfo{author}{B.~Crysmann}, \bibinfo{author}{W.~Packard},
\newblock \bibinfo{title}{Towards efficient {HPSG} generation for {G}erman, a non-configurational language.},
\newblock in: \bibinfo{booktitle}{COLING}, \bibinfo{year}{2012}, pp. \bibinfo{pages}{695--710}.
\bibitem[{Oepen and Flickinger(1998)}]{oepen1998towards}
\bibinfo{author}{S.~Oepen}, \bibinfo{author}{D.~Flickinger},
\newblock \bibinfo{title}{Towards systematic grammar profiling. test suite technology 10 years after},
\newblock \bibinfo{journal}{Computer Speech \& Language} \bibinfo{volume}{12} (\bibinfo{year}{1998}) \bibinfo{pages}{411--435}.
\bibitem[{Oepen(1999)}]{oepen1999incr}
\bibinfo{author}{S.~Oepen}, \bibinfo{title}{[incr tsdb ()] competence and performance laboratory. user and reference manual}, \bibinfo{year}{1999}.
\bibitem[{Packard(2014)}]{packard2014uw}
\bibinfo{author}{W.~Packard},
\newblock \bibinfo{title}{{UW-MRS}: Leveraging a deep grammar for robotic spatial commands},
\newblock \bibinfo{journal}{SemEval 2014}  (\bibinfo{year}{2014}) \bibinfo{pages}{812}.
\bibitem[{Copestake et~al.(2005)Copestake, Flickinger, Pollard, and Sag}]{copestake2005minimal}
\bibinfo{author}{A.~Copestake}, \bibinfo{author}{D.~Flickinger}, \bibinfo{author}{C.~Pollard}, \bibinfo{author}{I.~A. Sag},
\newblock \bibinfo{title}{Minimal recursion semantics: An introduction},
\newblock \bibinfo{journal}{Research on language and computation} \bibinfo{volume}{3} (\bibinfo{year}{2005}) \bibinfo{pages}{281--332}.
\bibitem[{Flickinger(2000)}]{flickinger2000building}
\bibinfo{author}{D.~Flickinger},
\newblock \bibinfo{title}{On building a more efficient grammar by exploiting types},
\newblock \bibinfo{journal}{Natural Language Engineering} \bibinfo{volume}{6} (\bibinfo{year}{2000}) \bibinfo{pages}{15--28}.
\bibitem[{Flickinger(2011)}]{Flickinger:11}
\bibinfo{author}{D.~Flickinger},
\newblock \bibinfo{title}{Accuracy v.\ robustness in grammar engineering},
\newblock in: \bibinfo{editor}{E.~M. Bender}, \bibinfo{editor}{J.~E. Arnold} (Eds.), \bibinfo{booktitle}{Language from a Cognitive Perspective: Grammar, Usage and Processing}, \bibinfo{publisher}{CSLI}, \bibinfo{address}{Stanford, CA}, \bibinfo{year}{2011}, pp. \bibinfo{pages}{31--50}.
\bibitem[{Carreras et~al.(2004)Carreras, Chao, Padr{\'o}, and Padr{\'o}}]{carreras2004freeling}
\bibinfo{author}{X.~Carreras}, \bibinfo{author}{I.~Chao}, \bibinfo{author}{L.~Padr{\'o}}, \bibinfo{author}{M.~Padr{\'o}},
\newblock \bibinfo{title}{Freeling: An open-source suite of language analyzers},
\newblock in: \bibinfo{booktitle}{Proceedings of the Fourth International Conference on Language Resources and Evaluation (LREC’04)}, \bibinfo{year}{2004}.
\bibitem[{Yamada et~al.(2020)Yamada, Davidson, Fern{\'a}ndez-Mira, Carando, Sagae, and S{\'a}nchez-Guti{\'e}rrez}]{yamada2020cows}
\bibinfo{author}{A.~Yamada}, \bibinfo{author}{S.~Davidson}, \bibinfo{author}{P.~Fern{\'a}ndez-Mira}, \bibinfo{author}{A.~Carando}, \bibinfo{author}{K.~Sagae}, \bibinfo{author}{C.~S{\'a}nchez-Guti{\'e}rrez},
\newblock \bibinfo{title}{{COWS-L2H}: A corpus of {S}panish learner writing},
\newblock \bibinfo{journal}{Research in Corpus Linguistics} \bibinfo{volume}{8} (\bibinfo{year}{2020}) \bibinfo{pages}{17--32}.
\bibitem[{Schneider and McCoy(1998)}]{schneider1998recognizing}
\bibinfo{author}{D.~Schneider}, \bibinfo{author}{K.~McCoy},
\newblock \bibinfo{title}{Recognizing syntactic errors in the writing of second language learners},
\newblock in: \bibinfo{booktitle}{ACL}, \bibinfo{year}{1998}, pp. \bibinfo{pages}{1198--1204}.
\bibitem[{Zamaraeva and Gómez-Rodríguez(2023)}]{zamaraeva2023revisiting}
\bibinfo{author}{O.~Zamaraeva}, \bibinfo{author}{C.~Gómez-Rodríguez}, \bibinfo{title}{Revisiting supertagging for {HPSG}}, \bibinfo{year}{2023}. \href{http://arxiv.org/abs/2309.07590}{{\tt arXiv:2309.07590}}.
\bibitem[{Chiruzzo and Wonsever(2020)}]{chiruzzo2020statistical}
\bibinfo{author}{L.~Chiruzzo}, \bibinfo{author}{D.~Wonsever},
\newblock \bibinfo{title}{Statistical deep parsing for spanish using neural networks},
\newblock in: \bibinfo{booktitle}{IWPT}, \bibinfo{year}{2020}, pp. \bibinfo{pages}{132--144}.
\bibitem[{Dridan(2013)}]{dridan2013ubertagging}
\bibinfo{author}{R.~Dridan},
\newblock \bibinfo{title}{Ubertagging: Joint segmentation and supertagging for english},
\newblock in: \bibinfo{booktitle}{EMNLP}, \bibinfo{year}{2013}, pp. \bibinfo{pages}{1201--1212}.

\end{thebibliography}



\end{document}